%% file: main.tex
\documentclass[letterpaper, 10 pt, conference]{ieeeconf}  

\IEEEoverridecommandlockouts                              

\usepackage{amsmath} 
\usepackage{amssymb}  
\usepackage{graphics} 
\usepackage{epsfig} 
\usepackage{mathptmx} 
\usepackage{times} 
\usepackage{caption}
\usepackage{booktabs}
\let\labelindent\relax
\usepackage{enumitem}
\usepackage{xcolor} 

\usepackage{amsthm}

\usepackage{hyperref}
\usepackage[
backend=biber,
style=numeric,
]{biblatex}
\newtheorem{theorem}{Theorem}

\usepackage{changepage}

\definecolor{barbie-pink}{HTML}{e0218a}

\newcommand\weblink{adaptive-compliance.github.io}

\hypersetup{
    colorlinks=true,
    linkcolor=blue,
    filecolor=magenta,
    urlcolor=blue,
    pdftitle={Your Paper Title},
    pdfpagemode=FullScreen,
}

\title{\bf
Adaptive Compliance Policy:\\
Learning Approximate Compliance for Diffusion Guided Control 
\vspace{-3mm} }

\author{Yifan Hou$^1$\quad  Zeyi Liu$^1$ \quad Cheng Chi$^1$ \quad  Eric Cousineau$^2$\quad  Naveen Kuppuswamy$^2$\quad  \\ Siyuan Feng$^2$\quad   Benjamin Burchfiel$^2$\quad   Shuran Song$^1$ 
\\
$^1$ Stanford University  $^2$ Toyota Research Institute \\
\url{\weblink}
}

\begin{document}


\twocolumn[{%
	\renewcommand\twocolumn[1][]{#1}%
	\maketitle
        \vspace{-4mm}
	\begin{center}
        \includegraphics[width=0.98\textwidth]{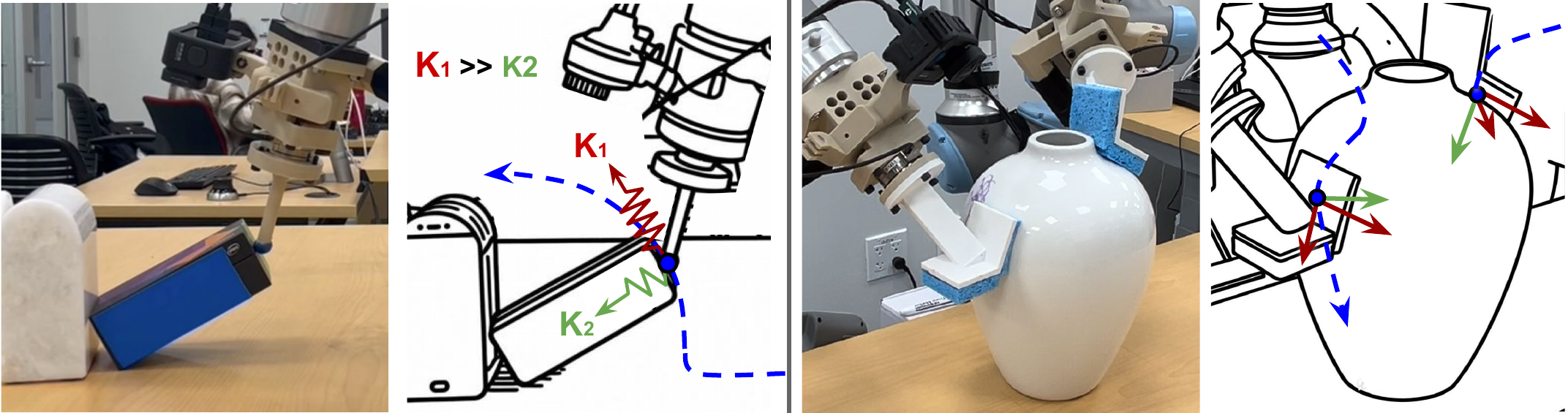}
        \captionof{figure}{ {\label{fig:teaser}\textbf{Compliance Requirements.}
        [Left] Flipping an item requires the robot to follow an arc trajectory (blue) while maintaining contact force. This demands low stiffness in pushing directions ($K_2$) and high stiffness elsewhere ($K_1$). [Right] Wiping a vase necessitates 3D compliance adjustments in both end-effectors to 1) hold the vase, 2) trace the marking, and 3) apply appropriate force without damage. Our algorithm aims to model these spatial-, temporal-, and task-dependent compliance requirements from human demonstration data.
        }}
	\end{center}
}]

\thispagestyle{empty}
\pagestyle{empty}

\begin{abstract}

Compliance plays a crucial role in manipulation, as it balances between the concurrent control of
position and force under uncertainties. 
Yet compliance is often overlooked by today's visuomotor policies that solely focus on position control. 
This paper introduces Adaptive Compliance Policy (ACP), a novel framework that learns to dynamically adjust system compliance both spatially and temporally for given manipulation tasks from human demonstrations, improving upon previous approaches that rely on pre-selected compliance parameters or assume uniform constant stiffness.
However, computing full compliance parameters from human demonstrations is an ill-defined problem. Instead, we estimate an approximate compliance profile with two useful properties: avoiding large contact forces and encouraging accurate tracking. 
Our approach enables robots to handle complex contact-rich manipulation tasks and achieves over 50\% performance improvement compared to state-of-the-art visuomotor policy methods. Project website with result videos:
\url{adaptive-compliance.github.io}.
\end{abstract}

\input{text/01_intro}

\input{text/02_relatedwork}

\input{text/03_method}

\input{text/04_experiments}

\section{CONCLUSIONS}
In this work, we show that Adaptive Compliance Policy is an effective visuomotor policy for compliant manipulation. Extensive real-world results show that our approach is able to extract useful compliance from human demonstration, and thereby significantly improve the success rate of two contact-rich manipulation tasks. 






\section*{ACKNOWLEDGMENT}
This work was supported in part by the Toyota Research Institute, NSF Award \#2143601, \#2037101, and \#2132519. We would like to thank Google and TRI for the UR5 robot hardware. The views and conclusions contained herein are those of the authors and should not be interpreted as necessarily representing the official policies, either expressed or implied, of the sponsors.


\printbibliography

\end{document}

%% file: text/01_intro.tex
\begin{figure*}[t]
    \centering \includegraphics[width=0.99\linewidth]{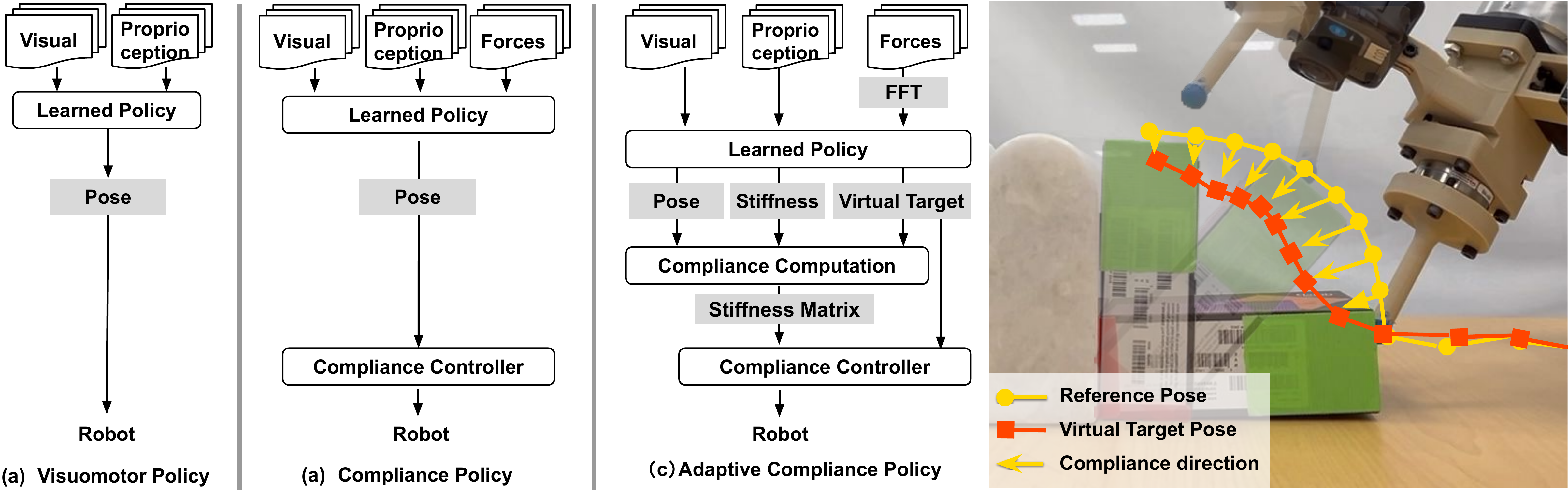}
    \caption{\textbf{Method Comparisons.} [LEFT] shows the comparison between a) a typical visuomotor policy \cite{chi2023diffusionpolicy}, b) a typical force-based compliant policy \cite{lee2019making}, and c) Adaptive Compliance Policy. [Right] Visualization of virtual target (orange sqaures) and reference poses (yellow circles) inferred by Adaptive Compliance Policy. The directional difference (orange arrows) between the virtual and reference poses encodes compliance direction.}
    \label{fig:method_overview}

    \vspace{-4mm}
\end{figure*}

\section{INTRODUCTION}
 Manipulation often requires the concurrent control of both position and force to achieve the desired outcome. 
This joint objective can be captured by the concept of mechanical compliance \cite{mason1981compliance,raibert1981hybrid,hogan1984impedance}, where a low compliance prioritizes position accuracy regardless of external forces, while a high compliance allows large position deviation in response to external forces, making the system ``soft'' during interaction. 

The desired compliance for a robotics system is not a static property; rather, it varies drastically depending on the task objectives and the system's state. For instance, consider the flipping task in Fig. \ref{fig:teaser}, the desired compliance:

\begin{itemize}[leftmargin=3.5mm]
    \item \textbf{Varies temporally}. For example, The system needs to be less compliant before contact to prioritize precise position tracking while becoming compliant upon contact.
    
    \item \textbf{Varies spatially}. For example, during the pivoting stage, the system should be compliant only in the pushing directions (i.e., $K_2$ direction) while maintaining high stiffness in other directions to follow the arc motion (e.g., low compliance in $K_1$ direction). 
    
    \item \textbf{Varies from task to task}. If we change to a different task, such as wiping a vase in Fig. \ref{fig:teaser} [Right], both temporal and spatial properties of the compliance will change in order to satisfy the unique 3D motion and force requirements.  
\end{itemize}

While the desired compliance can be obtained from optimization given physical measurements of the manipulation problem \cite{hou2019robust,hou2020manipulation}, 
it remains a challenge to obtain compliance parameters directly from human demonstrations. Compliance describes how force and motion variations in all directions are related, a typical demonstration trajectory does not contain all the information. Prior work often requires detailed known dynamics parameter \cite{grioli2011non} or multiple demonstration trajectories on the exact same task to statistically estimate one robot's compliance \cite{calinon2010learning,kronander2012online,duan2018learning,duan2019sequential}. These approaches cannot handle new scene configurations or unexpected perturbations. As a result, compliant policies either rely on pre-selected compliance parameters for the target tasks \cite{kamijo2024learning} or assume uniform constant stiffness across all directions \cite{lee2019making}.

In this work, we introduce \textbf{Adaptive Compliance Policy (ACP)}, a sensorimotor policy that learns to dynamically adjust the system compliance both \textit{spatially} and \textit{temporally} for a \textit{given manipulation task} from human demonstrations.

Specifically, the algorithm represents the compliance profile by an additional stiffness value and a virtual target pose in addition to the original reference pose (e.g., robot end-effector pose) predicted by the policy. The directional difference between the virtual and reference targets encodes the spatial distribution of stiffness. Finally, the predicted reference pose and stiffness can be executed by a standard low-level high-rate compliance controller to achieve robust and adaptive compliance behaviors.

Instead of estimating the exact human compliance, we derive a simple rule to obtain a useful compliance profile that guarantees the avoidance of large internal forces while encourages precise tracking, under mild assumptions about the tasks.
This simple rule allows us to approximate varying stiffness for every demonstration episode with different object variations and scene configurations.  
Then, by learning from a collection of demonstrations, the policy can summarize the typical compliance profile for a specific task and quickly adjust compliance based on visual and force feedback.   

We systematically evaluate the performance of our algorithm on two real world contact-rich manipulation tasks: object flipping and vase wiping. Our method achieves over 50\% increase in performance compared to state-of-the-art visuomotor policy methods.  
In summary, the main contribution of the paper includes: 
\begin{itemize}[leftmargin=6mm]
    \item Adaptive Compliance Policy formulation that is able to dynamically adjust compliance to maintain desired contact modes despite uncertainties and disturbances. 

    \item A kinesthetic teaching system that allows demonstrations with varying compliance profiles. 

    \item A method to compute spatial-, temporal-varying compliance labels from human demonstrations, making ACP training practical and scalable. 
\end{itemize}




    

%

%% file: text/02_relatedwork.tex
\section{RELATED WORK}

\noindent\textbf{Contact-rich manipulation with active compliance.}
There is a long history of work on utilizing robot compliance for robust contact-rich manipulation \cite{lozano1984automatic,uchiyama1988symmetric,sawasaki1991tumbling,hou2018fast}. Given modeling information such as contact geometry and friction, the stiffness that provides the maximum robustness can be computed efficiently \cite{hou2019robust,hou2021efficient,hou2020manipulation}. Although proved in many occasions, these methods typically require careful modeling work to setup. Our work utilizes similar mechanical modelings but derives a model-free method that is easier to scale.

\noindent\textbf{Learning compliance from Reinforcement Learning (RL).}
RL can learn compliance controllers by exploring force-motion variations \cite{portela2024learning,kalakrishnan2011learning, beltran2020learning,beltran2020variable, noseworthy2024forge, chang2022impedance,martin2019variable}. However, these policies need to be retrained for any scene variations. Moreover, most of existing work uses fixed-parameter low-level compliance controllers like impedance \cite{lee2019making} or admittance \cite{kohler2024symmetric} controllers, which lack robustness against disturbances.

\noindent\textbf{Learning compliance from human}
Human stiffness during manipulation can be estimated from sufficiently repeated same motions \cite{deng2016learning, duan2018learning,duan2019sequential,wang2020framework,yamane2023soft, zeng2021generalization}. The stiffness is either proportional to the force covariance \cite{duan2018learning,duan2019sequential} or inversely proportional to the position covariance \cite{calinon2010learning,kronander2012online}. These methods do not work for visuomotor policy learning since every demonstration is different. \cite{grioli2011non} proposed a stiffness estimator that works with a single demonstration. However, it requires perfect knowledge of human mass and damping, which is impractical. Prior work also attempted to give the demonstrator the ability to specify uniform stiffness \cite{kronander2013learning, kamijo2024learning}. 
A concurrent work \cite{liu2024forcemimic} computes compliance profiles based on heuristics from manually divided stages of tasks.




\noindent\textbf{Learning visuomotor policies with force feedback.} To effectively encode the temporal relations in the force/torque data, Prior work has explored many different encoding methods such as causal convolution \cite{lee2019making,}, TCN \cite{beltran2020variable}, LSTM \cite{ding2019transferable}, VAE \cite{aoyama2024few}, and self-attention \cite{kim2023training, kohler2024symmetric}. 
However, despite taking force as input, most of the existing visuomotor policies are still solely focused on predicting the position of the robot \cite{liu2024maniwav, li2022see} with uniform constant compliance \cite{lee2019making} that fails to capture the spatial- and temporal- variations of the compliance requirements for delicate manipulation task.



%% file: text/03_method.tex

\section{Method}
In this section, we introduce our pipeline, from collecting human demonstrations to designing the Adaptive Compliance Policy. We first introduce related physical concepts in  \S \ref{ref:sec:prelimnaries}, then describe a kinesthetic teaching system with compliance control   \S \ref{ref:sec:data}. 
Next, \S \ref{ref:sec:label} introduces how we annotate compliance labels for the human demonstration data. Finally, \S \ref{ref:sec:policy} explain the Adaptive Compliance Policy design.

\subsection{Preliminaries: Modeling \label{ref:sec:prelimnaries}}
\vspace{2mm}\noindent\textbf{Robot Compliance Modeling}.
Compliance expands the action space of a robot \cite{khatib1987unified,mason1981compliance}. Consider a $N$ dimensional system described by position $x\in \mathbb{R}^N$ and force $f\in \mathbb{R}^N$. Compliance is the elastic behavior between force and motion, which is typically modeled by a spring-mass-damper system:
\begin{equation}
\vspace{-2mm}
    f = M\ddot x + K(x - x_{ref}) + K_D\dot x.
    \label{eq:spring_mass_damper}
\end{equation}
The three terms on the right-hand side represent inertia force, spring force, and damping force, respectively. $x_{ref}$ is a reference position, at which the spring force is zero.
The compliant behavior is described by the inertia matrix $M\in R^{N\times N}$, stiffness matrix $K\in R^{N\times N}$ and damping matrix $K_D\in R^{N\times N}$, which can be user-specified if the compliance \ref{eq:spring_mass_damper} is implemented by control.
In other words, they can be added to the action space of a high level policy \cite{beltran2020learning,beltran2020variable,deng2016learning}, while a low level compliance controller implements the "virtual" stiffness/damping/inertia.
This is beneficial when the high level policy cannot be fast enough to exhibit compliance.

We use admittance control \cite{maples1986experiments} for our compliance controller, which takes in force feedback and outputs position targets. This works for our high accuracy position-controlled robot (UR5e). Robots with force control interface may also use impedance control \cite{hogan1984impedance}, or some hybrid force-motion control schemes \cite{mason1981compliance,raibert1981hybrid}.

\vspace{2mm}\noindent\textbf{Manipulation Modeling}.
Next we model how a manipulation system interacts with a robot, treating the robot as a black box. We make the following assumption:
\begin{quote} 
\begin{adjustwidth}{-2em}{-2em}
\textbf{Assumption I:} \textit{Contact force dominates. Other types of force, such as inertia force, friction and gravity, are negligible comparing with contact force.} 
\end{adjustwidth}
\end{quote}
This is not to be confused with the compliance control of the robot itself, which is fast and dynamic. We ensure Assumption I by avoiding fast robot motion and using lightweight objects. Consider a robot with $N$ degrees of freedom, making $n$ contacts with the environment. Denote $\lambda\in\mathbb{R}^n$ as the vector of contact normal forces, the \textit{Newton's Second Law} can be written as following with Assumption I: 
\begin{equation}
\vspace{-2mm}
    J^T\lambda = f,
    \label{eq: newtons_law}
\vspace{-1mm}
\end{equation}
where $J$ is the Contact Jacobian matrix that maps contact force into the robot generalized force space. While used in derivation, our method does not need to compute it. Denote $v\in\mathbb{R}^N$ as the generalized velocity vector, the Jacobian $J$ can also describe the velocity constraint imposed by contacts:
\begin{equation}
\vspace{-2mm}
    \label{eq:contact_constraints}
    Jv\ge0.
    \vspace{-2mm}
\end{equation}

\begin{figure}[t]
    \centering \includegraphics[width=0.98\linewidth]{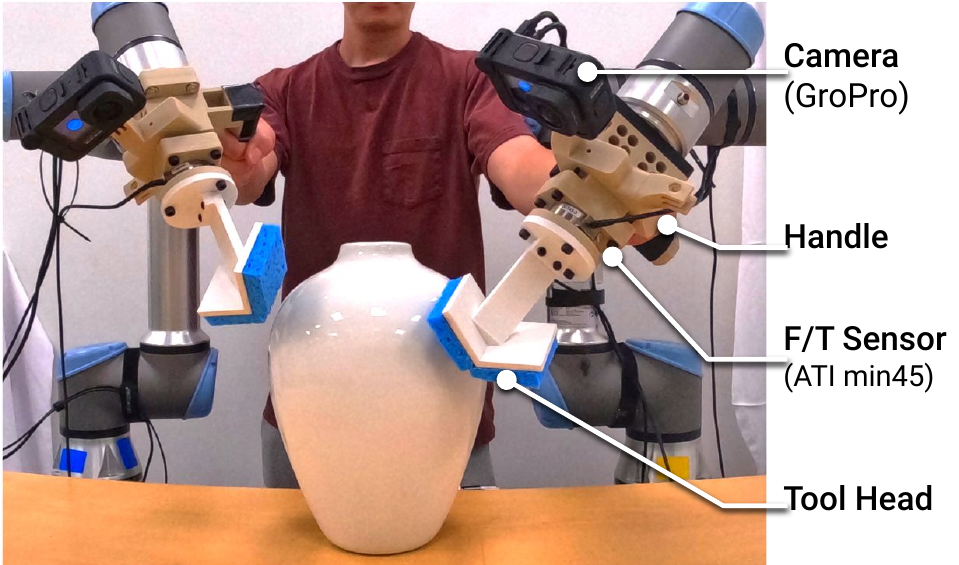}
    \caption{\textbf{Data Collection with Haptic Feedback.} We designed a kinesthetic teaching system with low-stiffness compliance that allows the operator to demonstrate variable compliance behavior with direct haptic feedback. }
    \label{fig:data}
    \vspace{-5mm}
\end{figure}

\subsection{Demonstration Collection \label{ref:sec:data}}
We choose kinesthetic teaching instead of teleoperation to collect human demonstrations in order for the operator to easily demonstrate variable compliance behavior under direct haptic feedback (see Fig. \ref{fig:data}). The setup for one arm includes one robot manipulator to provide accurate position feedback, one RGB camera to record visual information, and one force torque sensor mounted near the robot hand. 

During demonstration, we specify low stiffness, low damping and low mass for the robot compliance controller so the operator can move the robot freely.
Low damping and mass are achievable during demonstration because the human hand provides a natural external stabilization. During testing, we increase robot damping and virtual mass to maintain stability of the admittance controller. 
We also found it necessary to set the tool center point (TCP) near the handle, so the robot can rotate under external force in a way intuitive to the operator.

\subsection{Estimating Compliance from Demonstrations \label{ref:sec:label}}
Human uses varying stiffness during manipulation. High stiffness provides position accuracy under force disturbances. Low stiffness is also necessary, since the high stiffness controls impose velocity constraints that might conflict with the contact constraints defined in Eq. \ref{eq:contact_constraints}, during which a huge internal force may be generated \cite{hou2021efficient}.

However, it is often an ill-conditioned problem to estimate compliance parameters from a single human demonstration due to the lack of variations \cite{calinon2010learning}. Previous work on stiffness estimation \cite{grioli2011non} assumes perfect knowledge of damping and virtual mass and sufficient variations in force-motion signals. These requirements are not met in kinesthetic teaching, where the human hand changed the effective damping and mass of the robot hand, and the demonstration could have constant force or position for a period of time. 

To simplify the problem, we also use pre-selected values for mass and damping. Then instead of estimating the true human stiffness, we propose to find a stiffness matrix with the following properties:
\begin{itemize}
    \item It avoids huge internal forces in manipulation.
    \item It encourages accurate tracking of the desired motion.
\end{itemize}

\vspace{2mm}\noindent\textbf{C.1 Stiffness direction: }
We propose the following simple strategy to choose stiffness direction in the generalized space: \textit{Use a low stiffness $k_{low}$ in the direction of the force feedback, and a high stiffness $k_{high}$ in all other directions.} 

Next we explain why this simple rule works. 
From rigid body mechanics \cite{murray2017mathematical}, we know that the rows of Contact Jacobian $J$ represents the directions of contact normal forces, which forms a polyhedral convex cone in the generalized force space. We make two more assumptions:
\begin{quote} 
\begin{adjustwidth}{-2em}{-2em}
\textbf{Assumption II:} \textit{Nonzero contact force: all made contacts should have nonzero contact forces.} \\
\textbf{Assumption III:} \textit{No pinching contacts: the cone formed by rows of the Contact Jacobian $J$ is contained in its dual cone.}  
\end{adjustwidth}
\end{quote} 
Assumption II can be satisfied by making contacts clearly in demonstrations.
Assumption III means the contacts on the robot are not too restrictive. Fig. \ref{fig:pinching_example} shows some examples:

\begin{figure}[ht]
    \centering
    \includegraphics[width=0.9\linewidth]{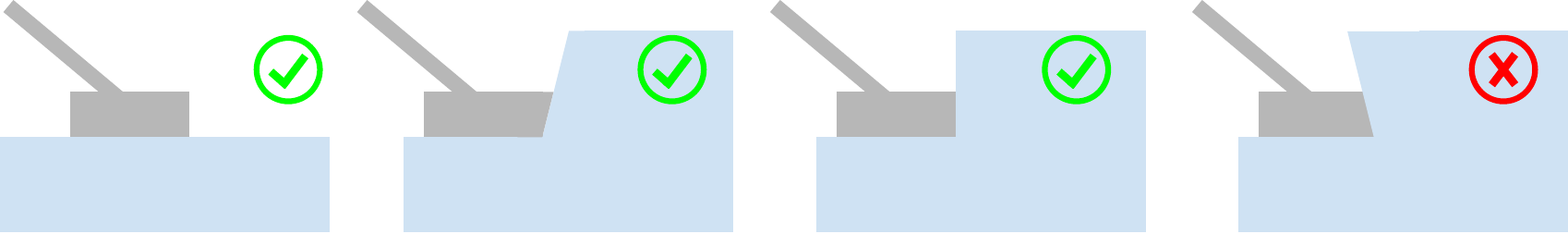}
    \caption{\textbf{Pinching Examples.} Grey shape represents a robot tool, blue shape represents a frictionless environment. First three examples are not pinching contact, the last one is.}
    \label{fig:pinching_example}
\end{figure}
We propose the following theorem under the assumptions: 
\begin{theorem}
For a robot under external contact described by Eq. \ref{eq: newtons_law}, there exists a solution $v$ that satisfies the contact constraint \ref{eq:contact_constraints} as long as it does not control its velocity in the direction of feedback force $f$ in the generalized space.  
\label{theorem:I}
\end{theorem}

\noindent\textbf{Proof.}
Not controlling velocity in the force direction means the velocity has a free component:
\begin{equation}
    v = v_0 + k f = v_0 + k J^T\lambda,
\end{equation}
where $k$ is an arbitrary scaling factor, $v_0$ denotes the components of generalized velocity in other directions. Due to Assumption II, the contact force $\lambda$ must have all positive components, $J^T\lambda$ represents a ray strictly inside the cone formed by rows of $J$. Assumption III says this cone is contained in its dual cone $\{x\in\mathbb{R}^N | Jx \ge 0\}$, so
\begin{equation}
    JJ^T\lambda > 0.
\end{equation}
Then $JV = JV_0 + k JJ^T\lambda > 0$ for large enough $k$.

Theorem \ref{theorem:I} shows that one-dimensional low stiffness control suffices to avoid constraint violation, so we can use high stiffness in other directions to improve position tracking.
Let $K_0\in \mathbb{R}^N$ be a diagonal matrix with $[k_{low}, k_{high}, ..., k_{high}]$ on its diagonal, and $S\in\mathbb{R}^N$ be a matrix whose columns form an orthonormal basis of $\mathbb{R}^N$ with its first column as $f/|f|$. The stiffness matrix can be written as:
\begin{equation}
    K = SK_0S^{-1}
    \label{eq: stiffness_matrix}
\end{equation}
We use $k_{high}$ in all directions when $|f|$ is small.

\vspace{2mm}\noindent\textbf{C.2 Stiffness Magnitude}
The high stiffness $k_{high}$ should support accurate position tracking in those directions, which can be set empirically. The low stiffness value should be zero, however, since the low stiffness direction is estimated from noisy force signal, we found it helpful to let the stiffness decrease continuously with the force magnitude: 
\begin{equation}
     k_{low}=\left\{\begin{matrix}
k_{max}, & |f|< f_{min}   \\
 k_{max} - (k_{max}-k_{min})\frac{|f|-f_{min}}{f_{max}-f_{min}}, &  f_{min}\le |f|\le f_{max} \\
k_{min}, & |f| > f_{max} \\
\end{matrix}\right.
\label{eq:stiffness_computation}
\end{equation}
where $k_{max}, k_{min}, f_{max}$ and $f_{min}$ are parameters determined by the hardware system.



\subsection{Adaptive Compliance Policy \label{ref:sec:policy}}
We formulate the policy as a diffusion process for both reference action and target stiffness.
The policy runs in a receding-horizon manner \cite{chi2023diffusionpolicy}, where an action trajectory is predicted using recent observations: 1) fisheye RGB images, 2) robot end-effector poses, and 3) force/torque data. 

\subsubsection{Inputs and Encoding}
We implement two encoding strategies for the force/torque data: 1) temporal encoding via causal convolution \cite{oord2016wavenet}, which helps capture causal relations from sequential data like force. We use force readings from the past 32 timesteps and pass them through a 5-layer causal convolution network, outputting a 768-dimensional vector; 2) FFT encoding, where we convert each dimension of the 6D force/torque readings into a 2D spectrogram. These spectrograms (6$\times$30$\times$17) are passed to a ResNet-18 model with a modified input channel of 6. A coordinate convolution layer \cite{liu2018intriguing} is added at the beginning to handle translational invariances. 
The images from the past two timesteps are resized to 224$\times$224 with random cropping then encoded using a CLIP-pretrained ViT-B/16 model \cite{dosovitskiy2020image}. 

Both image and force encodings are passed to a transformer encoder layer, using self-attention to learn adaptive visual-force representations. This representation is concatenated with robot end-effector poses from the past 3 timesteps and fed to the downstream diffusion policy head as a condition following \cite{chi2023diffusionpolicy}.

\subsubsection{Outputs and decoding}
Our policy output encodes the position target, the stiffness matrix, and the reference force in a 19-dimensional vector per robot arm:
\begin{itemize} 
    \item Reference pose: 9D pose vector following convention in \cite{chi2023diffusionpolicy}, where the last six elements are the top two rows of a rotation matrix;
    \item Virtual target pose: 9D pose representing the actual target for the low level compliance controller to track;
    \item A scalar value representing the stiffness magnitude in the low stiffness direction.
\end{itemize}
The virtual target pose is computed such that the robot will exert the reference force if it reaches the reference pose while tracking the virtual target with the given stiffness.
It essentially changes a force target into a position target. The benefit is to have a uniform target representation across different robots: an impedance controlled robot without FT sensor can also execute the virtual target.

During training, we first pass the whole episode of wrench data through a moving average filter with one second window size, then compute the stiffness from it using Eq. \ref{eq: stiffness_matrix}, and finally compute the virtual target following a 3D mechanical spring. The heavy wrench filtering has two benefits: 1) it makes the virtual target smooth; 2) it gives the action labels hindsight information about contacts about to be made, which is crucial for smooth contact engaging motions.

At inference time, the full stiffness matrix is reconstructed following Eq. \ref{eq: stiffness_matrix} by replacing the force direction with the direction from the reference pose to the virtual target. Finally both the stiffness matrix and the virtual target are sent to the low level compliance controller for execution.

%% file: text/04_experiments.tex
\section{Experiments}
We evaluate our method in two contact-rich manipulation tasks whose success depends on the maintenance of suitable contact modes. We use the UR5e robot with one GoPro RGB camera and one ATI mini-45 force torque sensor. The GoPro camera streams images at 60Hz, the robot receives Cartesian pose commands and send pose feedback at 500Hz, while the ATI sensor streams force-torque data at up to 7000Hz. 

We evaluate the following four policies, all trained on the same dataset with the same number of epochs:
\begin{itemize}[leftmargin=*]
    \item  ACP: Adaptive Compliance Policy, our approach; 
    \item  ACP w.o. FFT: same as ACP but with force encoded using temporal convolution \cite{oord2016wavenet, lee2019making} instead of FFT. 
    \item  Stiff policy: Diffusion policy \cite{chi2023diffusionpolicy,chi2024diffusionpolicy} with additional force input. Outputs target positions.
    \item  Compliant policy: Same as the [Stiff policy] except that the low level controller has a uniform stiffness $k=500$N/m. Relying on low level robot compliance is common in visuomotor policies \cite{lee2019making, wang2023mimicplay,zhu2023viola, kohler2024symmetric}.
\end{itemize}
Both ACP variations uses compliance in 3D translational space, i.e. $N=3$, though our method formulations works for 6D compliance too if needed. 
For all polices, we use two frames of RGB image and three frames of end-effector poses. The policy is not sensitive to these numbers when they are small. [ACP w.o. FFT] uses 32 frames of wrench data sampled at 250Hz, while all other three policies uses one second of data at 7000Hz.

\subsection{Task I: Item Flipping}
The task is to flip up an item with a point finger by pivoting it against a corner of a fixture (i.e., a wall), as exemplified in Fig. \ref{fig:method_overview}. The task has two challenges: 1) The finger needs to consistently maintain contact force during the flipping motion regardless of the item's shape, weight, and fixture locations; 2) Larger contact force will cause the fixture to slide on the floor, making it harder to maintain good contact. We trained each method on 230 episodes of demonstrations collected on 15 different items for 300 epochs.

\begin{figure}[ht]
    \centering
    \includegraphics[width=0.98\linewidth]{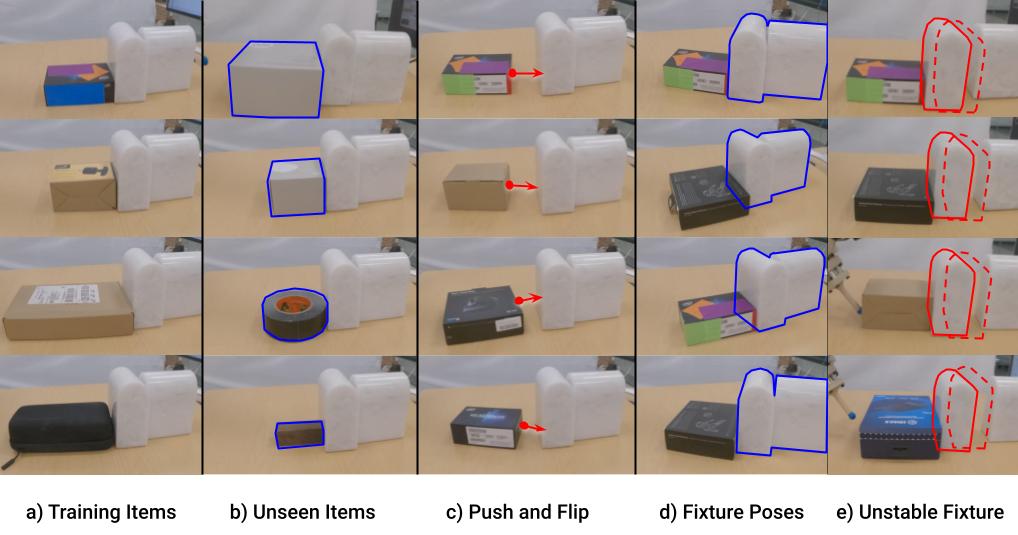}
    \caption{\textbf{Flipping Scenarios.} We test the policy under a variety of settings that require the policy to adapt to different and unseen object geometries (a,b), configuration (c,d) and react to unexpected perturbations caused by fixture movements (e).}
    \label{fig:flip_test}
    \vspace{-3mm}
\end{figure}

\vspace{2mm}\noindent\textbf{Test Scenarios. (Fig. \ref{fig:flip_test})} 
We ran 20 tests in each of the five scenarios below, making 100 tests per algorithm:
\begin{itemize}[leftmargin=*]
\item Training Items:  Items appeared in training data. 
\item Unseen Items: Items not seen in training.
\item Push\&Flip: Items start 5cm away from the fixture. This scenario requires the policy to first push the free-standing item towards the fixture then start the pivoting action.   
\item Varied Fixture Pose: Two different fixture poses. 
\item Unstable Fixture: Lighter fixture that causes unstable movements during the pivoting process that require the policy to quickly adapt. 
\end{itemize}

\begin{figure*}
    \includegraphics[width=0.98\linewidth]{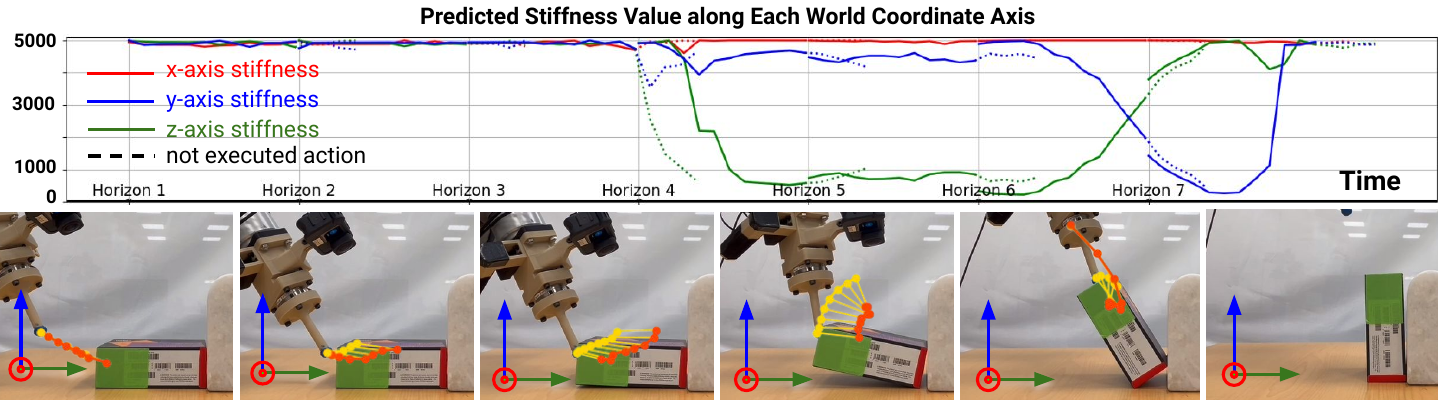}
    \label{fig:flip-result}
    \caption{\textbf{Flipping Results.} [Top] Predicted stiffness in the world frame.  The robot first decreases  X-axis stiffness while approaching from the same direction, then gradually shifts the compliance to Z-axis, aligning with the contact normal. [Bottom] Predicted reference pose (yellow dot),  virtual target poses (orange dot) and compliance direction (yellow line).}
    \vspace{-3mm}
\end{figure*}

\vspace{2mm}\noindent\textbf{Results.}
The success rate is shown in Tab. \ref{tb:flipping_result}. Success is defined as the item being rotated greater than 70 degrees. Typical failure cases include items falling off the finger, motion getting stuck, or triggering robot force violation.

\begin{table}[ht]
\centering
\caption{ Flipping-up Success Rates (\%)}
\label{tb:flipping_result}
\setlength{\tabcolsep}{4pt}
\begin{tabular}{l|ccccc|c}

\toprule 
  & Train & Unseen & Push & Fixture & Unstable & All\\ 
  & Items & Items & \&Flip & Pose & Fixture & \\ 
  \midrule
ACP                 & \textbf{90}   & 95    & 95   & \textbf{100} & \textbf{100} & \textbf{96}  \\  
ACP w.o. FFT        & \textbf{90}   & \textbf{100}    & \textbf{100}  & 95 & 90 & 95  \\  
Compliant Policy    & 80    & 15     & 15    & 5  & 0  & 23  \\ 
Stiff Policy        & 20     & 0    & 5   & 35   & 10   & 14   \\ 
\bottomrule  
\end{tabular}
\vspace{-5mm}
\end{table}

\vspace{2mm}\noindent\textbf{Findings. }
The two baselines [Compliant Policy] and [Stiff Policy] have a few successes when they can exploit the passive compliance in the system. They effectively applies a force when the predicted trajectory is in collision with a deformable item. When the item is rigid, or when the position uncertainty is not in a convenient direction, the baseline polices break the contacts and fail the task. On the contrary, both variations of ACP tolerates a large range of position uncertainties while maintain the needed contacts.

In this task, [ACP w.o FFT] has similar performance (95/100) as ACP (96/100), indicating that the force spectrum encoding is similarly useful to convolution in this task. This makes sense as the force signal does not contain much high frequency component in the flipping motion. 

\subsection{Task II: Vase Wiping}
The vase is randomly placed in front of the bimanual robot. The right upper side of the vase is marked by a random drawing using random colored dry markers. For this task, we equip each robot arm with a 3D-printed tool with two piece of kitchen sponges as wipers. The demonstrated motion uses the left arm to hold the vase while the right arm performs the wiping. We collected 200 demonstrations with various vase poses, marking shapes, and colors. Each demonstration includes one to five wipes to fully clean the markings.



\vspace{2mm}\noindent\textbf{Test Scenarios.} 
We compared with the same set of algorithm alternatives as in the Item Flipping task. For each algorithm, we ran a total of 16 tests in the following four scenarios: 
\begin{itemize}[leftmargin=*]
\item  Small Mark$\times5$: easier cases that need only one wipe.
\item  Large Mark$\times5$: require multiple wipes.  
\item  Perturbation before contact (PbC)$\times4$: move the vase right before the tool comes in contact with the vase to disturb the contact-engaging motion.     
\item  Perturbation after contact (PaC)$\times2$: move the vase after the tool is engaged to disturb the wiping motion. 
\end{itemize}

\vspace{2mm}\noindent\textbf{Results.}
The quantitative results are summarized in Tab. \ref{tb:wiping_result}. 
All policies demonstrated wiping behaviors. However, the effects of the wipes are different. We define success as the mark being cleaned within three wipes, where we consider a vase ``clean'' if the remaining marks are within 1cm$\times$1cm. 

\begin{figure}
    \includegraphics[width=\linewidth]{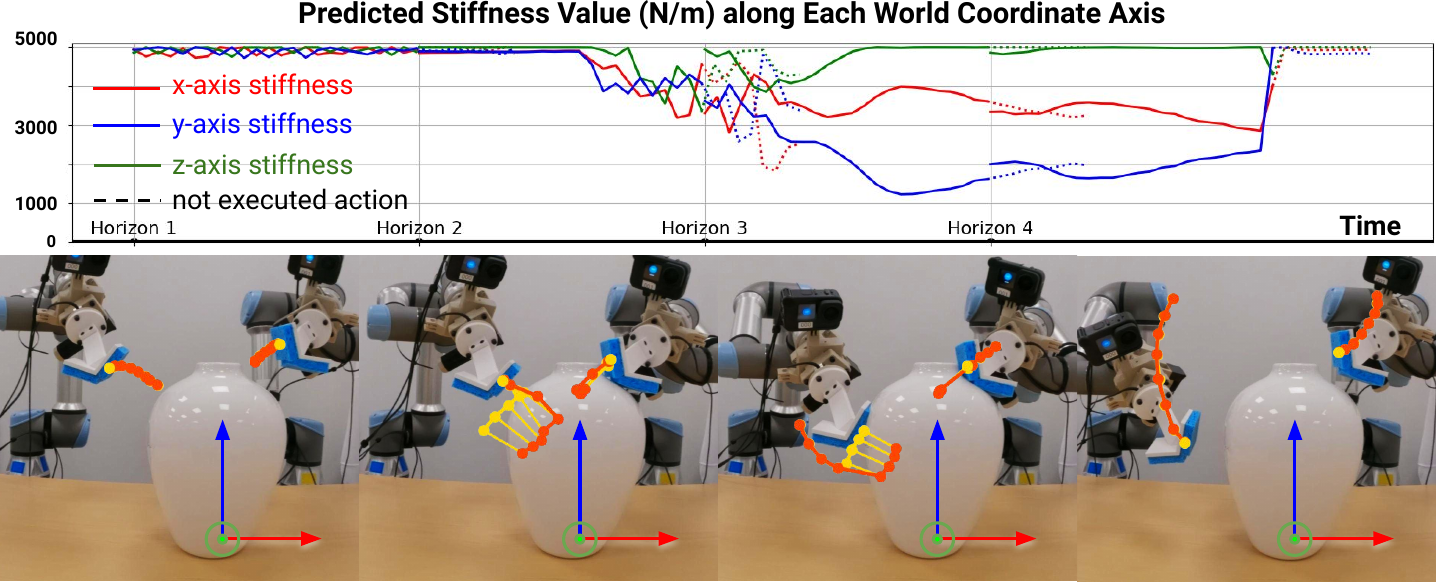}
    \label{fig:wipe-curves}
    \vspace{-5mm}
    \caption{ \textbf{Wiping Results.} [Top] Predicted stiffness of the wiping arm in world frame. [Bottom] Predicted reference (yellow dot) and virtual poses (orange dot). The Wiping arm's compliant direction roughly aligns with the contact normal. Despite some errors, the wiping succeeded as the virtual target pose still pulls arm towards the vase.} 
    \vspace{-5mm}
\end{figure}

\begin{table}[ht]
\centering
\caption{ Wiping  Success Rates (\%)}
\label{tb:wiping_result}
\setlength{\tabcolsep}{8.5pt}
\begin{tabular}{l|cccc|c}
\toprule 
  & Small & Large & PbC & PaC & All\\  
  \midrule
ACP                 & \textbf{100}   & \textbf{80}    & \textbf{100} & \textbf{100}  & \textbf{93.75} \\  
ACP w.o. FFT        & 100   & 60    & 75  & 100  & 81.25 \\  
Compliant Policy    & 60    & 20    & 25  & 100  & 43.75    \\ 
      
\bottomrule  
\end{tabular}
\vspace{-5mm}
\end{table}

\begin{figure}[ht]
    \centering    
    \includegraphics[width=\linewidth]{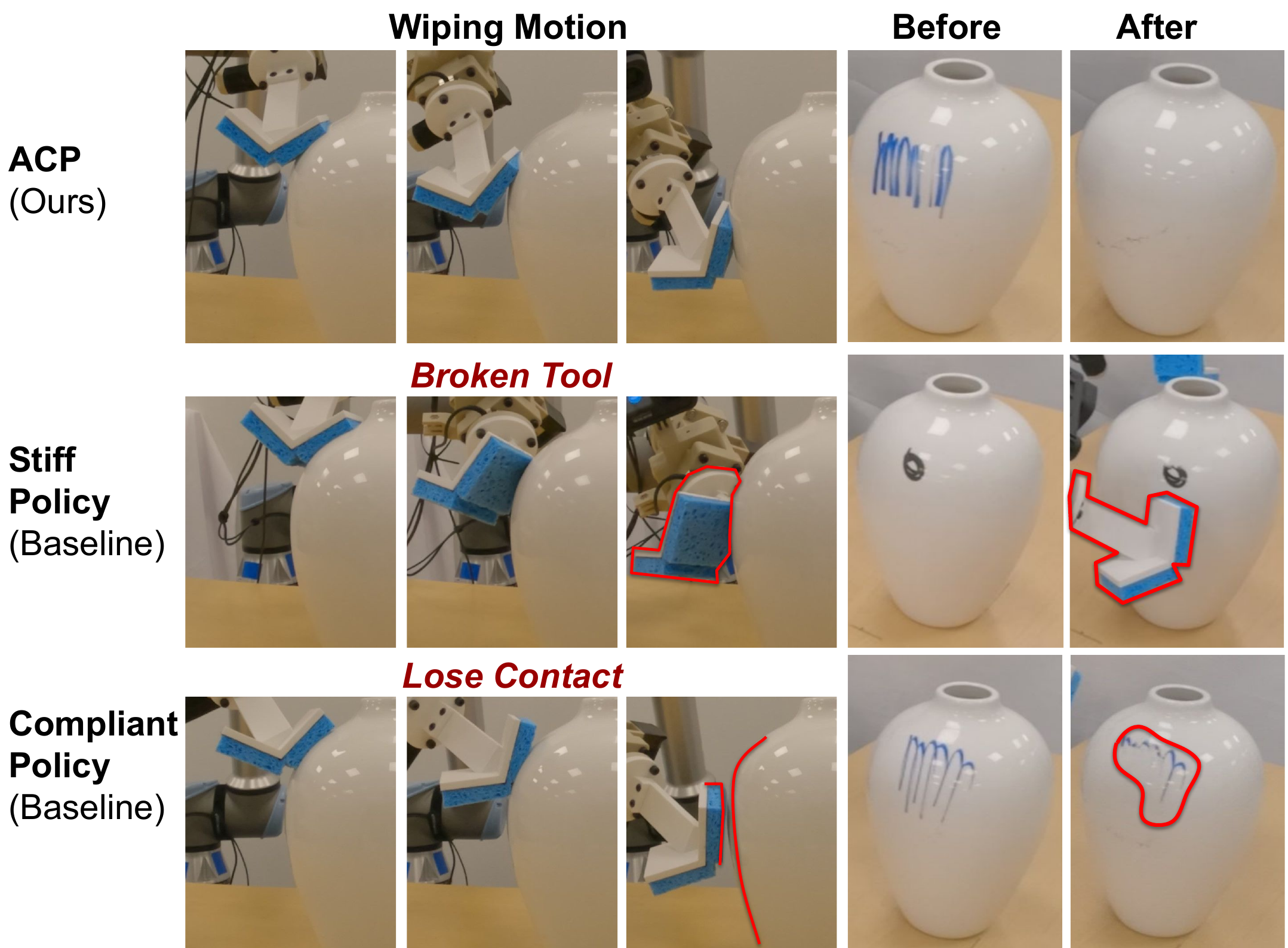}
    \vspace{-5mm}
    \caption{\textbf{Wiping Comparisons}. [Top] APC: maintains contact and follows desired trajectory. [Middle] Stiff Policy: Position noise causes excessive force that breaks the tool. [Bottom] Compliant Policy: Safe contact, but friction hinders wiping position accuracy and eventually loses contact. }
    \label{fig:wiping_motion}
    \vspace{-4mm}
\end{figure}

\vspace{2mm}\noindent\textbf{Findings. }
Compared with the [Stiff Policy], [ACP] safely engages and maintains contacts during the wipes for its compliance. [Stiff Policy]'s contact force magnitude varies greatly with the accuracy of the position action. While the first few tests were successful, the robot broke its tool during the fourth test, as shown in Fig. \ref{fig:wiping_motion}, middle row.

Compared with the [Compliant Policy], [ACP] maintains accurate tracking of the desired motion, as shown in Fig. \ref{fig:wiping_motion}, top row. The wiping motion of the [Compliant Policy] deviates from the position target because it is sensitive to the friction from the vase. As a result, the sliding motion is insufficient to make a high quality wipe. The comparison of wiping results is shown in Tab. \ref{tb:wiping_result}, which excludes the [Stiff Policy] since it broke the robot tool frequently.

Compared with [ACP w.o. FFT], our policy with the FFT encoding performs better and finishes the task with fewer wipes. We observed that the FFT encoding, when used together with RGB encoding via cross-attention, makes better decision on the next best wiping location.

We also compare with a policy that predicts force and position, which is less robust without a suitable compliance profile. See our website for details.